\documentclass[conference]{IEEEtran}
\usepackage[left=0.75in, right=0.75in, top=0.75in, bottom=0.75in]{geometry}
\usepackage[utf8]{inputenc}
\usepackage[T1]{fontenc}

\usepackage{cite}
\usepackage{float}
\usepackage{hyperref}
\usepackage{booktabs}
\usepackage{amsmath,amssymb,amsfonts, lipsum}
\usepackage{algorithmic}
\usepackage{graphicx}
\usepackage{textcomp}
\usepackage{xcolor}
\usepackage{optidef}

\usepackage[font=small,labelfont=bf,
   justification=justified,
   format=plain]{caption}

\begin{document}
\title{Masked Face Recognition for Secure Authentication}


\author{\IEEEauthorblockN{Aqeel Anwar\textsuperscript{1}, Arijit Raychowdhury\textsuperscript{2}}
\IEEEauthorblockA{\textit{Department of Electrical and Computer Engineering} \\
\textit{Georgia Institute of Technology, Atlanta, GA, USA}\\
\textit{aqeel.anwar@gatech.edu\textsuperscript{1}, arijit.raychowdhury@ece.gatech.edu\textsuperscript{2}}}
}




\maketitle
\thispagestyle{empty}
\pagestyle{plain}

\begin{abstract}

With the recent world-wide COVID-19 pandemic, using face masks have become an important part of our lives. People are encouraged to cover their faces when in public area to avoid the spread of infection. The use of these face masks has raised a serious question on the accuracy of the facial recognition system used for tracking school/office attendance and to unlock phones. Many organizations use facial recognition as a means of authentication and have already developed the necessary datasets in-house to be able to deploy such a system. Unfortunately, masked faces make it difficult to be detected and recognized, thereby threatening to make the in-house datasets invalid and making such facial recognition systems inoperable. This paper addresses a methodology to use the current facial datasets by augmenting it with tools that enable masked faces to be recognized with low false-positive rates and high overall accuracy, without requiring the user dataset to be recreated by taking new pictures for authentication.  We present an open-source tool, MaskTheFace to mask faces effectively creating a large dataset of masked faces. The dataset generated with this tool is then used towards training an effective facial recognition system with target accuracy for masked faces. We report an increase of $\sim38\%$ in the true positive rate for the Facenet system. We also test the accuracy of re-trained system on a custom real-world dataset MFR2 and report similar accuracy.
\end{abstract}

\flushbottom
\maketitle

\thispagestyle{empty}





\section{Introduction}
The world is currently under the onslaught of COVID-19. COVID-19 is an infectious disease caused by severe acute respiratory syndrome (SARS-CoV-2) \cite{paules2020coronavirus}. People can become infected by coming into close social contact with the infected person through respiratory droplets during coughing, sneezing and/or talking. Moreover, the virus can also be spread by touching a surface or object that has the virus on it, and then by touching your mouth, nose, or eyes. For now, we can protect ourselves by avoiding getting exposed to the virus. According to CDC the best way to avoid spreading or being infected with the disease is to practice social distancing and wearing face covering when in public areas \cite{CDC_guideline}. The two main prevention approaches are avoiding unnecessary contact and wearing face mask. Implementing these guidelines, seriously impacts the current security systems based on facial recognition that has already been put by several corporations and government organizations in place. Fingerprint or password-based security system, which involves contacting finger with sensor hence is not a good way to prevent the spread of disease making it unsafe. Face recognition-based security system however avoids unnecessary contact making it much safer than the former one. But such systems assume the that a picture of the entire face can be taken to perform recognition effectively. Widespread use of face masks thus renders the existing facial recognition systems in-efficient and they can make the entire infrastructure around facial recognition inoperable. 
Modern deep learning based face recognition systems have proven superior accuracy \cite{deng2019arcface, liu2016large, liu2017sphereface, liu2019fair, tuan2017regressing}. The accuracy of these systems depends on the nature of the available training images. Most of these systems assume access to un-occluded faces for recognition. This condition is fair when you can make sure that the system has access to the complete un-occluded face of the person being recognized. The system trained on such images learns to pay attention to important face features such as the eyes, nose, lips, face edges etc. But when these systems are presented a faced mask, the system fails to identify the person rendering the system unusable. We address this security problem with an effort to make the face recognition-based system reliable when presented with masked faces. The most important problem is the unavailability of the data for the system to be trained with. 

Our contributions are as follows
\begin{itemize}
    \item Open-source tool MaskTheFace to generate masked face dataset from face dataset with extended feature support
    \item Masked Faces in Real World for Face Recognition (MRF2) - A small dataset of aligned masked faces in real world
    \item Using MaskTheFace to retrain existing facial recognition system to improve accuracy.
\end{itemize}




\begin{figure*}[ht]
\centering
  \includegraphics[width=\linewidth]{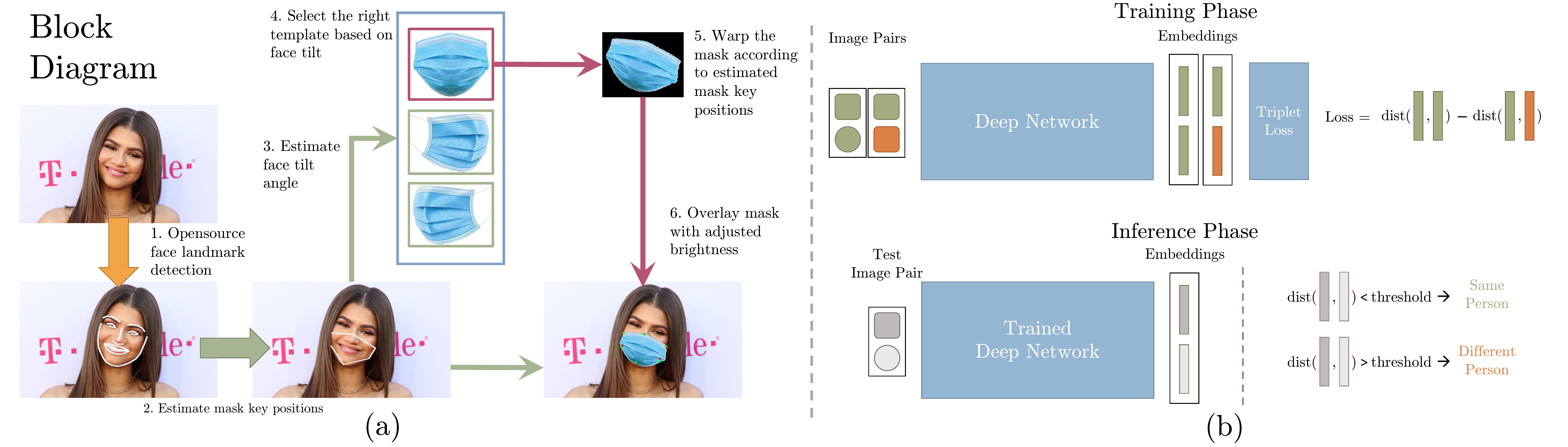}
  \caption{System level block diagrams (a) MaskTheFace tool warps the mask template selected by user based on the key face landmark positions of the face. (b) Training and Inference phase diagram of Facenet system - Facenet maps image pairs to embedding and compares the distance between them to mark the input pair belonging to same identity or different}
  \label{fig:block}
\end{figure*}

Due to lack of the masked face dataset, we propose using simulated masked face to train a deep network for face recognition through our open-source masking tool MaskTheFace. MaskTheFace is computer vision-based script to mask faces in images. It uses a dlib \cite{dlib} based face landmarks detector to identify the face tilt and six key features of the face necessary for applying mask. Based on the face tilt, corresponding mask template is selected from the library of mask. The template mask is then transformed based on the six key features to fit perfectly on the face. The complete block diagram can be seen in Fig. \ref{fig:block} (a).
MaskTheFace provides about $100$ different mask variations to select from. It is difficult to collect mask dataset under various conditions. MaskTheFace can be used to convert any existing face dataset to masked-face dataset. It identifies all the faces within an image, and applies the user selected masks to them taking into account various limitations such as face angle, mask fit, lighting conditions etc. A single image, or entire directory of images can be used as input to code.

Following are the features of the proposed system:
\begin{itemize}
    \item \textbf{Support for multiple mask types:} MaskTheFace provides support for 5 different mask types. Moreover, new custom face masks can be easily added by the user.
     \item \textbf{Support for mask variations:} MaskTheFace provides 24 existing patterns that can be applied to mask types above to create more variations of the existing mask types. Moreover, users can easily add custom patterns and colors following the guidelines provided.
    \item \textbf{Support for both single and multi-face images:} MaskTheFace can apply masks to all the faces in the images without any restriction of one face per image.
    \item \textbf{Wide face angle coverage:} Each face mask has multiple templates based on angle hence covering a wide range of face tilts.
    \item \textbf{Bulk masking on dataset:} Face datasets can be easily converted to masked face datasets by providing the folder path
\end{itemize}

There are many useful face datasets available such as CelebA \cite{liu2015faceattributes}, CASIA webfaces \cite{yi2014learning}, Labeled faces in the wild (LFW) \cite{huang2008labeled} and VGGFace2 \cite{Cao18}, to name a few, for the application of face detection/recognition. MaskTheFace can be used to convert these existing datasets into masked-face dataset which can then be used to train an efficient deep network for the underlying application.




\section{Experimentation}

In this paper, we point out the significant degradation in performance when a state-of-the-art face recognition system is presented masked images for recognition and then propose a solution that regains the degraded performance. This section covers the details on the selected face recognition system, training data and the performance metrics used to evaluate the trained network.

\subsection{Face recognition system:}
We select the state-of-the-art deep network-based face recognition system Facenet \cite{schroff2015facenet}. Facenet creates unified embeddings of the faces and then compares the faces in the embedding space to carry out decision making. The complete block diagram for Facenet based face recognition system can be seen in Fig. \ref{fig:block} (b). During the training phase, multiple image pairs are provided to the network. The network maps these image pairs to embedding vectors and calculates triplet loss \cite{weinberger2009distance}. The triplets consist of two image pairs of same and different identities. Triplet loss aims to separate the pair of same identity (positive pair) from the pair of different identities (negative pair) by a distance margin.
\begin{equation}
\left\|f\left(x_{i}^{a}\right)-f\left(x_{i}^{p}\right)\right\|_{2}^{2}+\alpha<\left\|f\left(x_{i}^{a}\right)-f\left(x_{i}^{n}\right)\right\|_{2}^{2}
\end{equation}

\begin{equation*}
\forall\left(f\left(x_{i}^{a}\right), f\left(x_{i}^{p}\right), f\left(x_{i}^{n}\right)\right) \in \mathcal{T}
\end{equation*}

where $f(.)$ is the underlying deep network to be trained, $x^a, x^p, x^n$ are the embedding of the anchor, corresponding positive and negative image. $\alpha$ is an L2 distance margin that is imposed between positive and negative pairs and $\mathcal{T}$ is the set of all possible triplets under consideration. Online triplet mining method is used to generate the triplets on the go to minimize the triplet loss. 

\begin{table*}[ht]
\small
\centering
\begin{tabular}{@{}cccccc@{}}
\toprule
\textbf{Dataset}     & \textbf{Type}&\textbf{\# Identities} & \textbf{\# Images} & \textbf{\# Avg. images/identity} & \textbf{\# Testing pair} \\ \midrule
VGGFace2-mini & Simulated &8,631                            &  362,502                     &   42                                    & -          \\ \midrule
VGGFace2-mini-SM1 & Simulated& 8,631                            & 697,084                    & 80.77                                         & -          \\ \midrule
LFW-SM (combined)   & Simulated    & 5,749                            & 64,973                       & 11.3                                       & 29,235     \\ \midrule
MFR2    & Real-world  & 53                              & 269                          & 5                                          & 848                              \\ \bottomrule
\end{tabular}
\caption{Summary of the four datasets used for training and inference. The first two datasets are used for training, while the last two are used for testing the trained network}

\label{tab:dataset}
\end{table*}

\subsection{Training data:}
For the purpose of training facenet, we select VGGFace2 \cite{Cao18} which is a large-scale face dataset that contains about 3 million images of $9131$ identities, with an average of $\sim362$ images per identity varying in pose, age, ethnicity and illumination. From VGGFace2 we create a subset, VGGFace2-mini, by randomly sampling 42 images per identity. This dataset contains the un-masked images of the identities. From VGGFace2-mini, we further generate VGGFace2-mini-SM dataset by applying randomly selected masks (surgical-green, surgical-blue, N95, cloth) to each image nearly doubling the dataset size. Table \ref{tab:dataset} summarizes these datasets.


For comparison purpose, we train two different Facenet networks, one with VGGFace2-mini (no-mask network) and the other with VGGFace2-mini-SM (mask network). Comparing the performance of two networks trained on datasets of different sizes is normally and un-fair comparison since the network trained on the larger dataset had access to more information hence having a tendency to make better decision. In our case, the extra images in VGGFace2-mini-SM are generated from the original images and they won't add any extra information for training no-mask network, hence the comparison is fair enough. We use Inception-resnet v1 \cite{szegedy2017inception} as the deep network $f(.)$ to map images to their embedding. An embedding size of 512 was selected. Each of the two networks were trained from scratch. 

\subsection{Performance Metrics:}
To analyze the performance of the trained networks, we use the following metrics as used in \cite{schroff2015facenet}
\begin{itemize}
    \item \textbf{Max Accuracy (\%):} The maximum accuracy of the network in terms of identifying the test input image pairs as -ve or +ve.
    
    \item \textbf{ACC @ FAR=0.1\% (\%):} The accuracy of the network in terms of identifying the test input image pairs as -ve or +ve at the selected threshold for which the false acceptance rate is 0.1\%. 
    
    \item \textbf{TPR @ FAR=0.1\% (\%):} The true positive rate (\%age of the time that the +ve input image pairs were identified +ve) of the network at the selected threshold for which the false acceptance rate is 0.1\%. 
\end{itemize}
To summarize
\begin{equation*}\label{eq:acc}
Max Accuracy = \frac{TP+TN}{TP+TN+FP+FN}
\end{equation*}

\begin{equation*}\label{eq:accatfar}
\begin{multlined}
ACC@FAR 0.1\% = \frac{TP+TN}{TP+TN+FP+FN}  \\
 s.t. ~~ \frac{FP}{TP+FP}=0.001 ~~~~~~~~~
 \end{multlined}
\end{equation*}

\begin{equation*}\label{eq:tpratfar}
\begin{multlined}
TPR@FAR 0.1\% = \frac{TP}{TP+FN} ~~~~~~~~~~~~~~~~~ \\
 s.t. ~~ \frac{FP}{TP+FP}=0.001 ~~~~~~~~~
 \end{multlined}
\end{equation*}

where TP, TN, FP and FN are True positives, true negatives, false positives and false negatives respectively.


\section{Results}
In this section we compare the performance of the trained no-mask and mask networks on the following test datasets based on performance metrics mentioned in the previous section
\subsection{LFW-SM - Dataset with simulated masks}
Labelled Faces in the Wild (LFW) dataset \cite{huang2008labeled} is a standard benchmark dataset used to evaluate the performance of face recognition systems. It contains $5,749$ identities with a total of $13,233$ images. As mentioned previously, the goal of masked face recognition is to accurately recognize identities both with and without the masks on. For this purpose, we use the following variations of the LFW dataset to draw effective comparisons.

\begin{itemize}
    \item \textbf{LFW:} Original unmasked LFW dataset.
    \item \textbf{LFW-SM-SG:} LFW dataset with the surgical-green simulated mask applied.
    \item\textbf{LFW-SM-SB:} LFW dataset with the surgical-blue simulated mask applied.
    \item \textbf{LFW-SM-N95:} LFW dataset with the N95 simulated mask applied.
    \item \textbf{LFW-SM-Cloth:} LFW dataset with the cloth simulated mask applied.
    \item \textbf{LFW-SM-Mixed:} LFW dataset with one randomly selected mask applied to each image.
\end{itemize}

LFW-SM dataset only contains images with the simulated mask applied on them. The purpose of these variations is to provide a detailed performance analysis. 

\begin{figure*}[ht]
\centering
  \includegraphics[width=\linewidth]{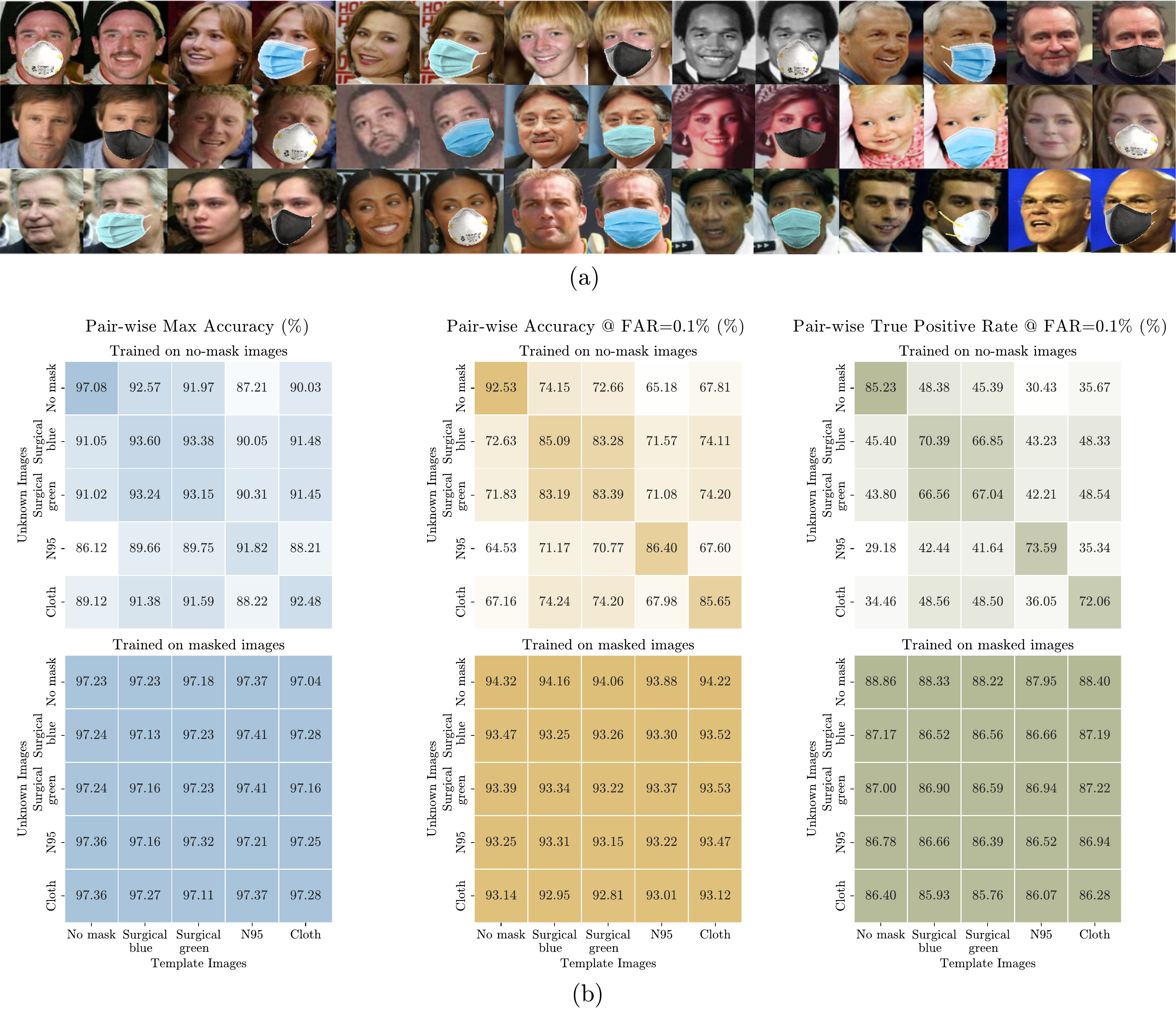}
  \caption{(a) LFW-SM-Mixed Dataset - Simulated mask images with $5,749$ Identities, $64,973$ Images and $29,235$ pairs. Mask types surgical, cloth and N95 is used to generate simulated masked dataset from the LFW dataset. (b) Detailed performance metrics of the no-mask and mask networks tested on LFW-SM dataset. Each performance metric is plotted as heat-map for visual comparison for 25 different possible image pairs. On the horizontal axis are the template images, while on the vertical axis we have unknown image of the test pairs. The darker the cell color, the better the performance metric}
  \label{fig:lfw_masked}
\end{figure*}

The facenet network takes input a pair of images (template-image, unknown-image) comparing the identity of the person in unknown image to the template image. This comparison is carried out in the embedded space based on the distance between the two. The selection of the L2 distance threshold is carried out on 9 splits of the underlying test dataset and the 10th split is used to report the performance metrics, similar to leave-one-out cross validation approach. The images in the test pairs are selected across the datasets to evaluate the robustness of the network. We use the standard protocol for unrestricted, labeled outside data as mentioned \cite{huang2014labeled} to evaluate $6,000$ image pairs. The no-mask network is trained on the un-masked faces. The L2 distance threshold is therefore calculated on the LFW original unmasked dataset. This L2 distance threshold is found by averaging out the optimum threshold for the 9 out of 10 splits and then is kept constant across all the variations of image pairs. The selected optimal threshold is $2.5$ for maximum accuracy and $1.67$ for FAR=0.1\%. The mask network is trained on a combination of masked and un-masked faces. The L2 distance threshold is therefore calculated on the LFW-SM-mixed dataset. Similar to the no-mask network, the L2 distance threshold is found by averaging out the optimum threshold for the 9 out of 10 splits and then is kept constant across all the variations of image pairs. The selected optimal threshold is $2.62$ for maximum accuracy and $2.01$ for FAR=0.1\%. Fig. \ref{fig:lfw_masked} (b) reports the performance metrics when the network was evaluated on LFW-SM dataset.

\begin{figure*}[ht]
\centering
  \includegraphics[width=\linewidth]{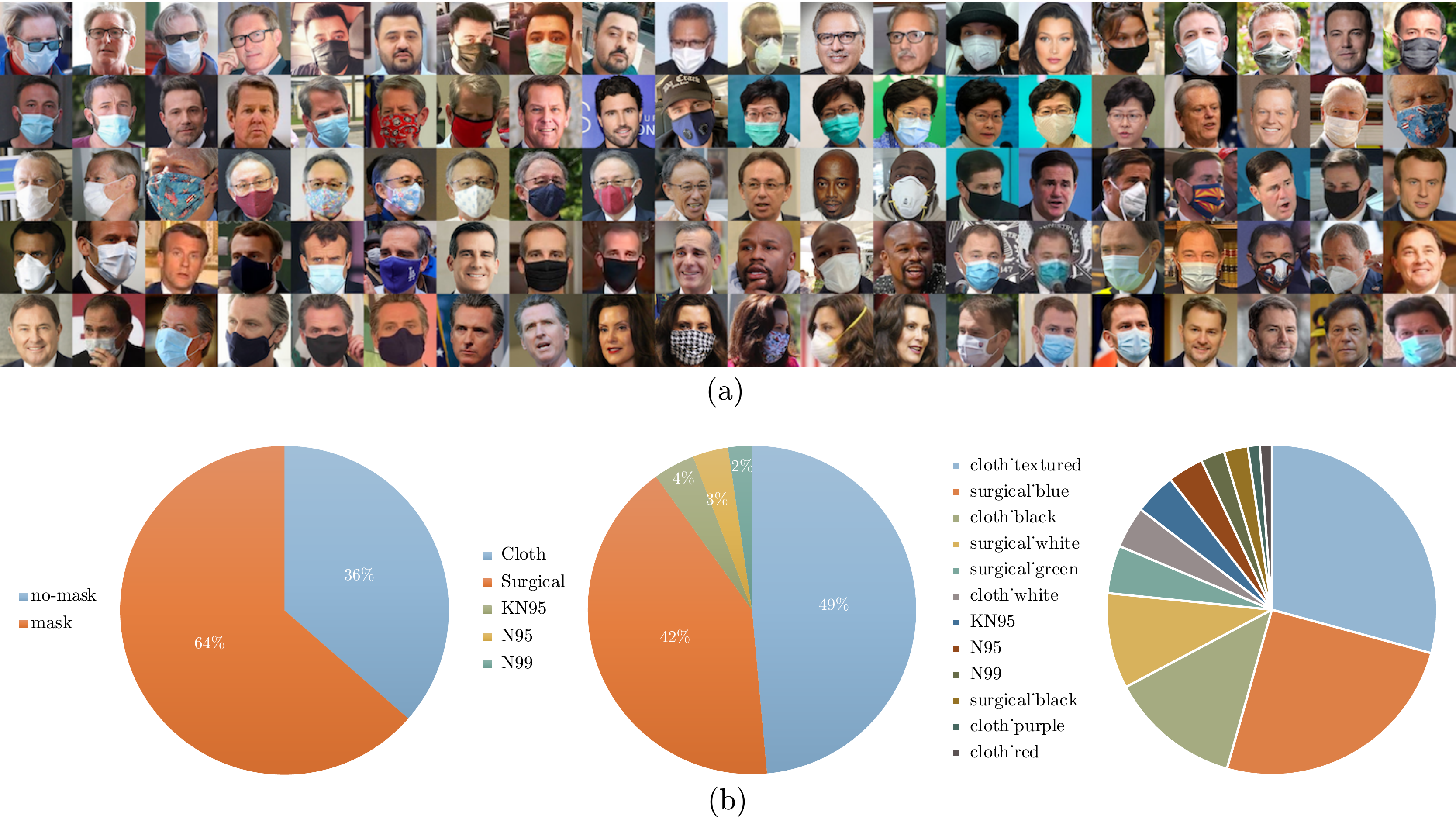}
  \caption{Summary of the MFR2 dataset. (a) Sample images from the MFR2 dataset containing masked face images of various politicians and celebrities wearing different types of masks. (b) The distribution of the MFR2 dataset for various mask types (legends are in decreasing order).}
  \label{fig:mfr2}
\end{figure*}

\begin{figure*}[ht]
\centering
  \includegraphics[width=\linewidth]{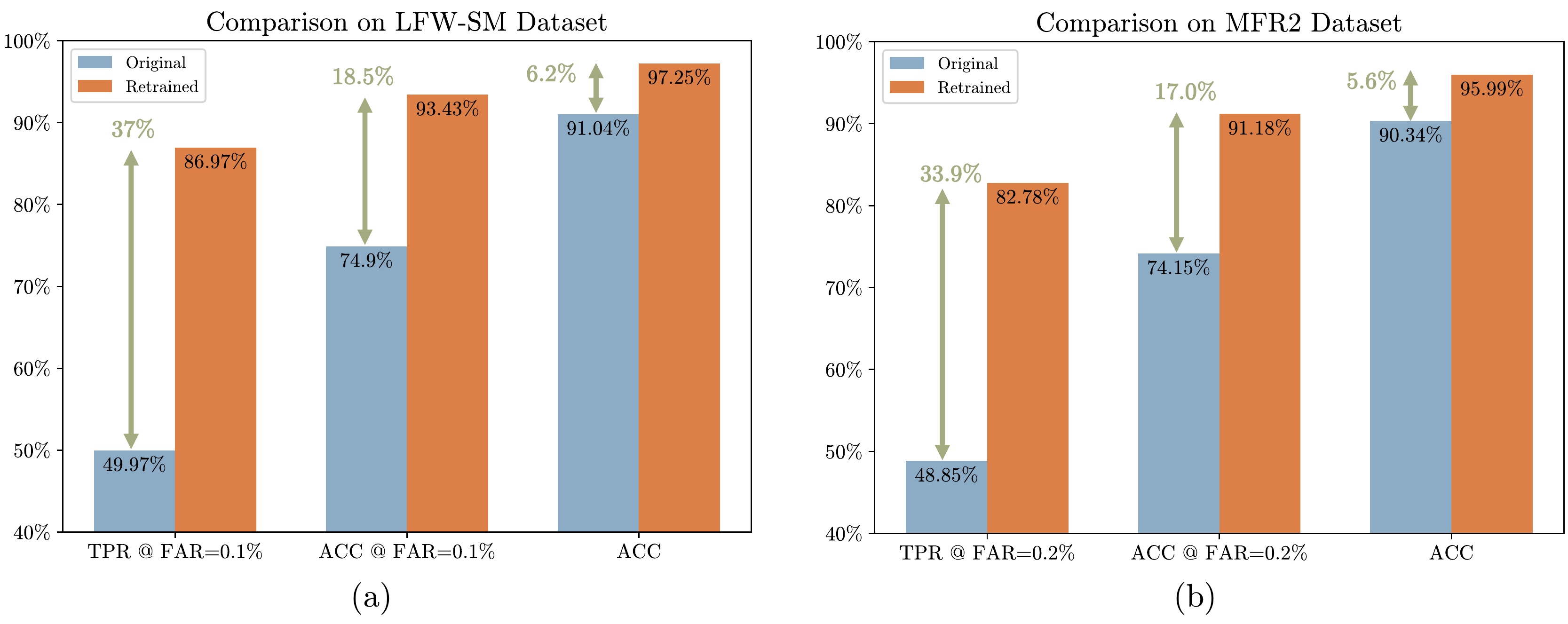}
  \caption{Comparison of performance metrics for the simulated LFW-SM and real-world MFR2 dataset for the no-mask and mask networks. Both the datasets result in similar performance making the solution robust enough to be implemented on real images.}
  \label{fig:comparison}
\end{figure*}

\subsection{MFR2 - Dataset with real masks}
 In the previous section we showed the network performance on the LFW Dataset with simulated masks applied through MaskTheFace tool. In this section we study the performance of the system trained on simulated masks but tested on the real-world masked faces. To the best of our knowledge there is no diverse enough, pre-processed, ready-to-use real-world aligned masked face dataset of identities. \cite{wang2020masked} has, by far, the largest face recognition dataset with 24,771 images. But the dataset faces are not consistent or aligned making it a little harder to be used. Moreover, the masks on the face makes it really hard to find and align faces for the purpose of face recognition. Also, the dataset lacks unmasked faces of the identities and is not diverse enough in terms of ethnicity. Therefore, we decided to create a rather smaller but pre-processed dataset of real-world masked faces consisting of celebrities and politicians. This dataset is not a replacement of the Wuhan dataset, but simply an approach for us to show the effectiveness of the simulated masks dataset on the real-world masked images. Masked faces in real world for face recognition (MFR2) is a small dataset with 53 identities of celebrities and politicians with a total of 269 images that are collected from the internet. Each identity has on average of 5 images. The dataset contains both masked and unmasked faces of the identities. The dataset is processed in terms of face alignment and image dimensions. Each image has a dimension of $(160\times160\times3)$. Images from the MFR2 dataset and the distribution of different mask types can be seen in Fig. \ref{fig:mfr2}. In the future we plan on expanding this dataset to contain more identities and images.
For the purpose of network performance evaluation, we consider a total of $848$ image pairs from MFR2 ($424$ positive pairs, and $424$ negative pairs). The pairs are evaluated through the network trained on VGGFace2-SM dataset. The overcome the gap between the simulated mask images and real-world mask images, the distance threshold is re-calculated for MFR2 using data splits as was done for LFW-SM. The performance of the network when evaluated on MFR2 is plotted in Fig. \ref{fig:comparison}. We compare the performance of no-mask network and mask network. Since this dataset is smaller in size than the LFW dataset, a FAR of 0.1\% means all the 894 pairs being predicted correct ($1/894 = 0.0012 > 0.001$). We relax this constraint to 0.22\% ($2/894 = 0.0022$) giving the network room for 2 wrong predictions.

\section{Discussion}
\subsection{LFW-SM Dataset}
Fig. \ref{fig:lfw_masked} (b) reports the detailed performance metrics of the two networks mentioned above on the LFW-SM dataset. Each performance metric is plotted as heat-map for visual comparison for 25 different possible image pairs. On the horizontal axis are the template images, while on the vertical axis we have unknown image of the test pairs. The darker the cell color, the better the performance metric. Fig. \ref{fig:lfw_masked} (b) left reports the pair-wise maximum accuracy of the no-mask (top) and mask (bottom) network. It can be seen that the accuracy is best ($\sim 97\%$) for the pairs when both the template and unknown image belongs to the un-masked dataset, while for all the other combinations the accuracy varies between $86$ to $93\%$. This might not seem a significant drop in the accuracy, but when things are kept in perspective, we realize that this performance metric might not be the best one for comparison. The accuracy is the average accuracy of the binary decision whether the pairs belonged to same identity or not. A better metric is to maintain the false acceptance rate to below a certain threshold (0.1\%) \cite{schroff2015facenet}. False acceptance rate (FAR) is the ratio of image pairs of different identities which was predicted same identity by the network over all the negative pairs. We want this metric to be as small as possible (ideally zero). An acceptable value of 0.1\% was selected for the FAR and the corresponding L2 distance threshold was calculated. The image pairs were then evaluated for their accuracy and true positive rate (TPR) at this threshold.

Fig. \ref{fig:lfw_masked} (b) center reports the accuracy at @FAR=0.1\%. It can be seen that the accuracy dropped to 76 to 85\% when masked faces were presented to the network. The worst accuracy is for the combination where the template image is no-mask and the unknown image is N95 masked-image. Furthermore, if we look at the TPR@FAR=0.1\% in Fig. \ref{fig:lfw_masked} (b) right, we see that the no-mask trained network drops the TPR from $\sim 86\%$ to $\sim30\%$ in the worst case scenario. This significant drop in the TPR makes the no-mask trained network highly vulnerable to prediction errors rendering the face recognition system in-efficient. Fig \ref{fig:lfw_masked} (b) also plot the results for the mask network (on the right). It can be seen that the network trained on the MaskTheFace generated dataset performs almost equally well for all the cross-dataset pair-wise combination. In fact, in some cases it slightly outperforms the no-mask trained network. The reason for this slight improvement could be the masked images which act as regularizers avoiding the network to overfit to VGGface2-mini and hence performing slightly better on the unseen test dataset. Fig. \ref{fig:comparison} (a) compares the average performs of the no-mask and mask networks across all the combinations for the considered performance metrics. It can be seen that the mask network significantly outperforms the no-mask network across all the three performance metrics. 

\subsection{MFR2 dataset}
 The resulting graphs show similar behavior to what was observed for the case of LFW-SM dataset. It is important to note that MFR2 dataset contain masks (such as cloth masks in different textures and colors - Fig. \ref{fig:mfr2}) that the trained network had never seen before. But the network was still able to achieve comparable accuracy on MFR2 as compared to LFW-SM. We observe an improvement of $\sim34\%$, $\sim17\%$ and $\sim6\%$ in TPR@FAR=0.2\%, Accuracy @ FAR=0.2\% and maximum accuracy respectively. Hence, we conclude that the network trained with the help of masked images generated from MaskTheFace is robust enough and performs reasonably well ($2$ to $4\%$ decrease in performance) when tested on real-world images.

\begin{figure*}[ht]
\centering
  \includegraphics[width=\linewidth]{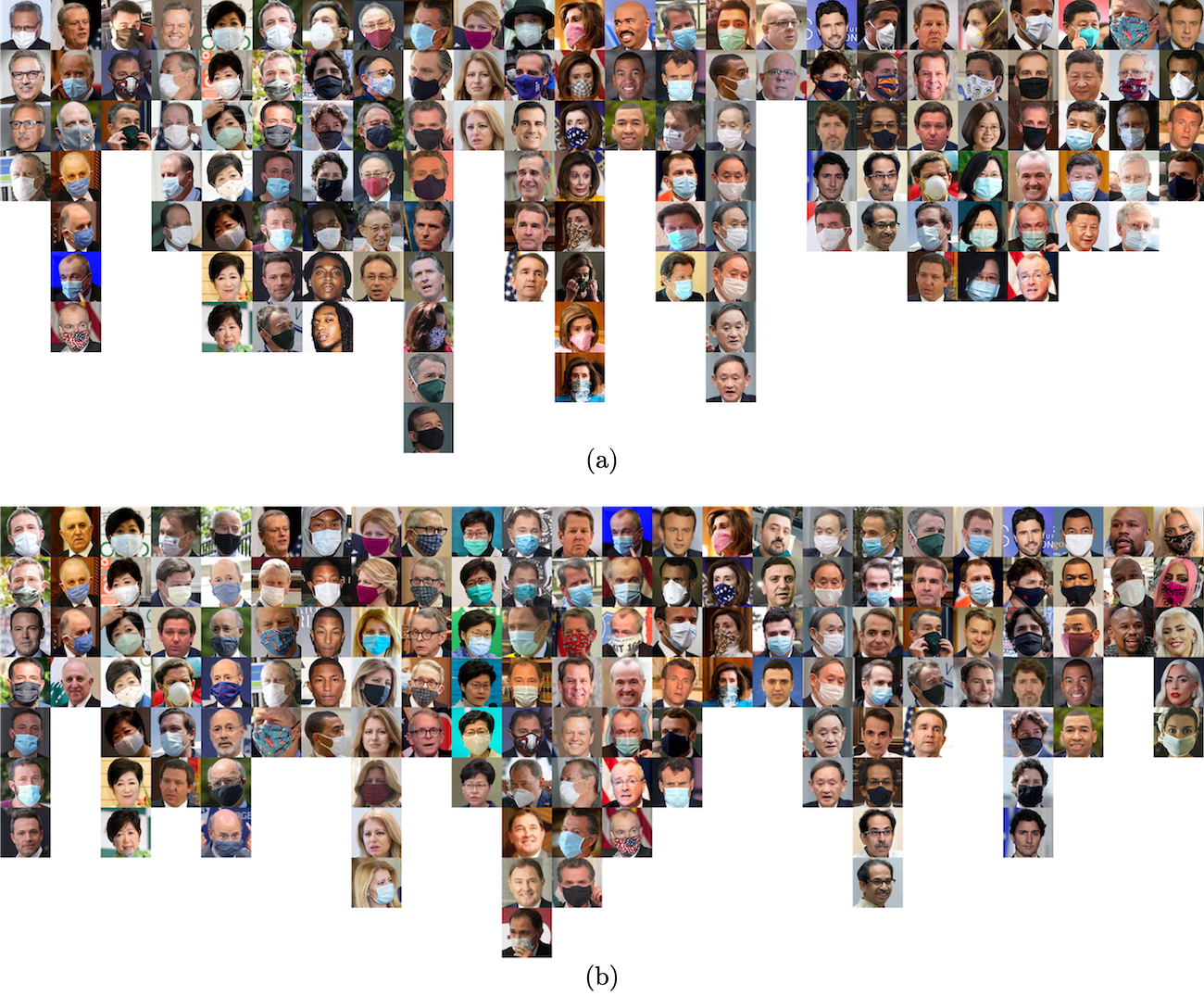}
  \caption{Clustering MFR2 Identities based on the embedding (a) MFR2 dataset clustered based on the no-mask network. (b) MFR2 datset clustered based on the mask network. There can be seen a few discrepancies in the clustering of images based on no-mask network, such as images mapped to wrong identities, same identities mapped to different clusters, different identities mapped to same cluster and different identities mapped into one cluster based on mask nature. On the other hand, we only see a single type of discrepancy (image mapped to wrong identity) in the clustering of images based on mask network, and that too for only a few clusters, while all the other images are mapped into correct clusters.
}
  \label{fig:cluster}
\end{figure*}

\subsection{Clustering MFR2 dataset}
Facenet created embedding of the images can be used to cluster faces into identities. We tried to cluster and the MRF2 dataset into clusters based on the embedding created by the no-mask and mask network. Fig. \ref{fig:cluster} (a) shows the MRF2 dataset clustered into groups based on the no-mask network. We only show 24 out of 53 clusters for better view. It can be seen that the no-mask network has some tendency of grouping images of same identities into one cluster (and hence a good accuracy from previous section). But there can be seen a few discrepancies in the clustering such as images mapped to wrong identities, same identities mapped to different clusters, different identities mapped to same cluster and different identities mapped into one cluster based on mask nature and hence a reduced accuracy and tpr @FAR=0.2\%. 
Fig. \ref{fig:cluster} (b) shows the clustering of MFR2 dataset based on the embedding created by the mask network. We only see single type of discrepancy (image mapped to wrong identity) and that too for only a few clusters, while all the other images are mapped into correct clusters.

\section{Methods}
\subsection{Dataset generation using MaskTheFace}
Faces dataset were converted into masked face dataset using MaskTheFace. For each image in the dataset, a mask was selected from cloth, surgical-green, surgical-blue and N95 uniformly at random. Moreover, the original un-masked image was made a part of the dataset alongside with masked images. This was done to make sure that the network trained, performs equally well on both the masked and un-masked images. 

\subsection{Training Details}
The VGGFace2 dataset was aligned and cropped using MTCNN \cite{zhang2016joint} to get a tight bound on the faces in the images. This gets rid of the redundant background in the image putting more focus on the face. The training of the facenet system was carried out on a GTX1080 workstation for 100 epochs. An equally spaced three-step learning rate with values 0.05, 0.005 and 0.0005 was used. Training each network took $\sim42$ hours.

\section{Conclusion}
In this paper we addressed the issue of recognizing masked faces through existing face recognition systems with reliable accuracy. We present an open-source tool, MaskTheFace which can be used to mask faces. This results in the creation of a large dataset of masked faces. The dataset generated with this tool can then used towards training an effective facial recognition system with target accuracy for masked faces. Using MaskTheFace, we report an increase of $\sim38\%$ in the true positive rate for the existing Facenet system for both masked and un-masked faces. The accuracy of re-trained system was also tested on a custom real-world dataset MFR2 and reported similar accuracy, hence being able to extend out to real life masked faces.

\section{Code availability}
MaskTheFace tool and the MFR2 dataset is available for use to other researchers at the following link \href{https://github.com/aqeelanwar/MaskTheFace}{https://github.com/aqeelanwar/MaskTheFace}. 

\section{Acknowledgements} 
This work was supported in part by C-BRIC, one of six centers in JUMP, a Semiconductor Research Corporation (SRC) program sponsored by DARPA.

\bibliographystyle{ieeetran}
\bibliography{main}

\end{document}